\documentclass{article}

     \usepackage[preprint, nonatbib]{neurips_2020}

\usepackage[utf8]{inputenc} % allow utf-8 input
\usepackage[T1]{fontenc}    % use 8-bit T1 fonts
\usepackage{hyperref}       % hyperlinks
\usepackage{url}            % simple URL typesetting
\usepackage{booktabs}       % professional-quality tables
\usepackage{amsfonts}       % blackboard math symbols
\usepackage{nicefrac}       % compact symbols for 1/2, etc.
\usepackage{microtype}      % microtypography

\usepackage{graphicx} % more modern
\usepackage{amsfonts} % for checkmark cmd
\usepackage{amsmath,soul}
\usepackage{amssymb}
\usepackage[capitalize]{cleveref} %cref
\newcommand{\unit}[1]{\ensuremath{\, \mathrm{#1}}} % non-italic units

\usepackage{dsfont} % mathds
\usepackage[font=small,skip=0pt]{subcaption} %subfigure
\usepackage{balance}
\usepackage[font=small]{caption}
\usepackage{bm} % bold italic vectors
\usepackage{wrapfig}
\usepackage{float} % To place figure on top right as float

\usepackage{makecell}

\usepackage{color}
\usepackage[svgnames]{xcolor} \definecolor{DarkGreen}{rgb}{0,0.5,0}
\definecolor{DarkRed}{rgb}{0.75,0,0}

\usepackage{tikz,mathtools}
\crefformat{equation}{(#2#1#3)}
\Crefformat{equation}{Equation~(#2#1#3)}
\Crefname{equation}{Equation}{Equations}

\usetikzlibrary{shapes,positioning,automata,arrows,fit,backgrounds,calc}
\tikzstyle{block} = [draw, fill=blue!20, rectangle,minimum height=1em,
minimum width=2em] \tikzstyle{sum} = [draw, fill=blue!20, circle, node
distance=1cm] \tikzstyle{input} = [coordinate] \tikzstyle{output} =
[coordinate] \tikzstyle{pinstyle} = [pin edge={to-,thin,black}]
\usetikzlibrary{trees} \usetikzlibrary{decorations.pathmorphing}
\usetikzlibrary{decorations.markings}
\definecolor{darkgreen}{rgb}{0,0.5,0}
\definecolor{darkred}{rgb}{220,20,60}

\newcommand{\cmmnt}[1]{\ignorespaces}

\newcommand{\bit}{\begin{itemize}}
\newcommand{\ei}{\end{itemize}}

\makeatletter
\renewcommand\paragraph{\@startsection{subsubsection}{4}{\z@}%
{0.25ex \@plus.5ex \@minus.2ex}%
{-.15em}%
{\normalfont\normalsize\itshape}}
\makeatother

\usepackage[authormarkuptext=name,addedmarkup=bf,authormarkupposition=left]{changes}
\definechangesauthor[name={X.~X.}, color={orange}]{xx}
\definechangesauthor[name={B.~L.}, color={MediumSeaGreen}]{bl}
\definechangesauthor[name={A.~P.}, color={DeepPink}]{ap}
\title{Physics-informed GANs for coastal flood visualization}
\author{%
  Bj\"orn L\"utjens$^{*,1}$, Brandon Leshchinskiy$^{*,1}$, Christian Requena-Mesa$^{*,2}$, \\
  \textbf{Farrukh Chishtie$^{*,3}$, Natalia D\'iaz-Rodr\'iguez$^{*,4}$, Oc\'eane Boulais$^{*,5}$, Aaron Pi\~na$^{6}$} \\ \textbf{Dava Newman$^1$, Alexander Lavin$^7$, Yarin Gal$^8$, Chedy Ra\"issi$^9$} \\ \thanks{\texttt{lutjens@mit.edu}, $^1$Human Systems Laboratory, Massachusetts Institute of Technology, $^2$Max Planck Institute for Biogeochemistry, $^3$Spatial Informatics Group, 2529 Yolanda Ct., Pleasanton, CA 94566, USA$^4$Autonomous Systems and Robotics Lab, Inria Flowers, ENSTA Institut Polytechnique Paris, $^5$Responsive Environments Research Group, MIT Media Lab, $^6$NASA Headquarters, $^7$Latent Sciences, $^8$Oxford Applied and Theoretical Machine Learning Group, University of Oxford, $^9$Ubisoft}$\; $ equal contribution 
}

\begin{document}

\maketitle

\begin{abstract}
As climate change increases the intensity of natural disasters, society needs better tools for adaptation. Floods, for example, are the most frequent natural disaster, but during hurricanes the area is largely covered by clouds and emergency managers must rely on nonintuitive flood visualizations for mission planning.
To assist these emergency managers, we have created a deep learning pipeline that generates visual satellite images of current and future coastal flooding.
We advanced a state-of-the-art GAN called pix2pixHD, such that it produces imagery that is physically-consistent with the output of an expert-validated storm surge model (NOAA SLOSH).
By evaluating the imagery relative to physics-based flood maps, we find that our proposed framework outperforms baseline models in both physical-consistency and photorealism.
While this work focused on the visualization of coastal floods, we envision the creation of a global visualization of how climate change will shape our earth\footnote[10]{Code and data will be made available at \url{github.com/repo}; an interactive demo at \url{trillium.tech/eie}.}.
\vspace{-.1in}
\end{abstract}

%\begin{IEEEkeywords}
%Keywords: Physics-informed deep learning, coastal flooding, climate change, generative adversarial networks, visualization
%\end{IEEEkeywords}

\begin{figure}[htbp!]
  \centering
  \begin{subfigure}{.8\columnwidth}
      \centering
      \includegraphics [trim=0 0 0 0, clip, width=1.\textwidth, angle = 0]{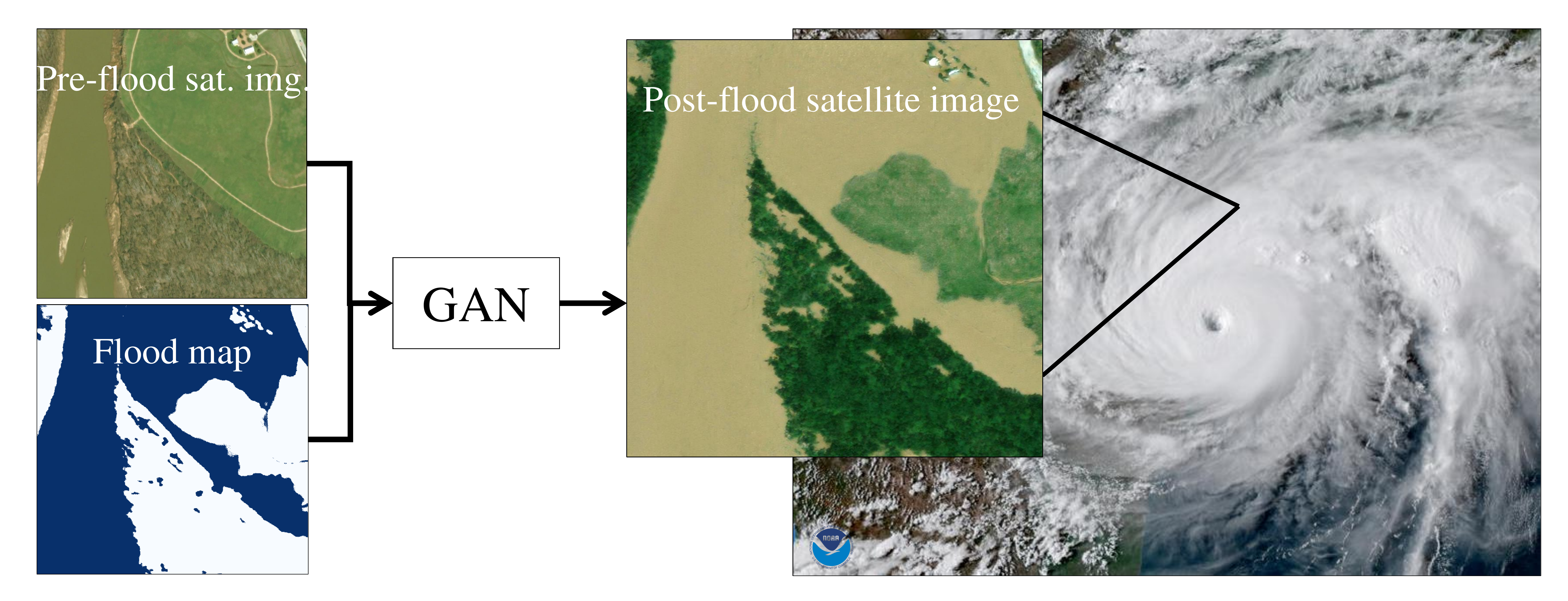}
      \vspace{-.25in}
  \end{subfigure}
\caption[Model Architecture]{\textbf{We have generated the first physically-consistent satellite imagery of coastal flood events (mid).} During hurricanes, such as Harvey in Houston, TX, the area is largely covered by clouds and emergency responders must rely on nonintuitive flood maps, inferred from models or radar, for mission planning (bottom-right, \cite{NoaaSlosh_20}). We propose a novel GAN-based pipeline to convert the flood maps into photorealistic post-flood images, enabling emergency managers to intuitively understand flood impacts.
} \label{fig:model_architecture} 
\vspace{-.2in}
\end{figure}

\section{Introduction}\label{sec:intro}
\begin{figure}[htbp!]%t]
  \centering
  \begin{subfigure}{1.\columnwidth}
      \centering
      \includegraphics [trim=0in 0.0in 0in .0in, clip, width=1.\textwidth, angle=0]{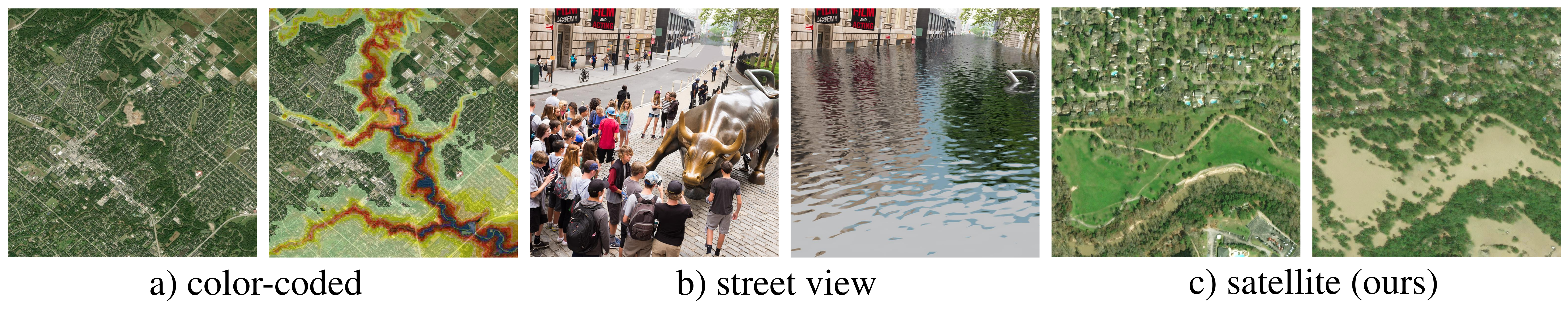}%trim=0.5in 0.1in 1.8in .4in
      \vspace{-.1in}
      %\caption{Before training}
  \end{subfigure}
\caption[related works]{\textbf{Existing coastal flood visualizations are either informative or intuitive, but not both.} State-of-the-art flood visualizations include either color-coded geospatial rasters (a), which can be nonintuitive to interpret, or photorealistic street view imagery (b), which is intuitive, but too narrow to provide an overview for city-scale climate resilience planning. Our visualizations (c) are both intuitive and informative, while maintaining physical validity. (Image sources:~\cite{ NAIP_2019, NoaaSlosh_20, Strauss_2015, Strauss_2015, Gupta_2019}, ours)}\label{fig:related_works} 
\end{figure}

As our climate changes, natural disasters become more intense~\cite{IPCC_2018}. Floods are the most frequent weather-related disaster~\cite{Cred_2015} and already cost the U.S. $3.7\unit{B \; USD}$ per year; this damage will only grow over the next several decades~\cite{NoaaNcei_2020, IPCC_2018}. 
Today, emergency managers and local decision-makers rely on visualizations to understand and communicate flood risks (e.g., building damage)~\cite{NOAA_2020}. 
Shortly after a coastal flood, however, clouds cover the affected area and before a coastal flood no RGB satellite imagery exists to plan flood  response or resilience strategies. 
Existing visualizations are limited to informative overviews (e.g., color-coded maps~\cite{NOAA_2020, NoaaSlosh_20, ClimateCentral_18}) or intuitive (i.e., photorealistic) street-view imagery~\cite{Schmidt_2019, Strauss_2015}.
Expert interviews, however, suggest that color-coded maps (displayed in~\cref{fig:related_works}a) are non-intuitive and can complicate communication among decision makers~\cite{Radovan_2020}.
Street-view imagery (displayed in~\cref{fig:related_works}b), on the other hand, offers intuitive understanding of flood damage, but remains too local for city-wide planning~\cite{Radovan_2020}. %mission-planning.
To assist with both climate resilience and disaster response planning, we propose the first deep learning pipeline that generates satellite imagery of coastal floods, creating visualizations that are both intuitive and informative. 

% Our work is more scalable than Schmidt_2019, because paired images exist 
% Our work is more scalable than physics-based renderings, because it does not require explicit modeling of each scene.
%\blumargin{*}{todo: mention we're not forecasting floods}

\begin{figure}[t]
  \centering
  \begin{subfigure}{1.\columnwidth}
      \centering
      \includegraphics [trim=0 0 0 0, clip, width=1.\textwidth, angle = 0]{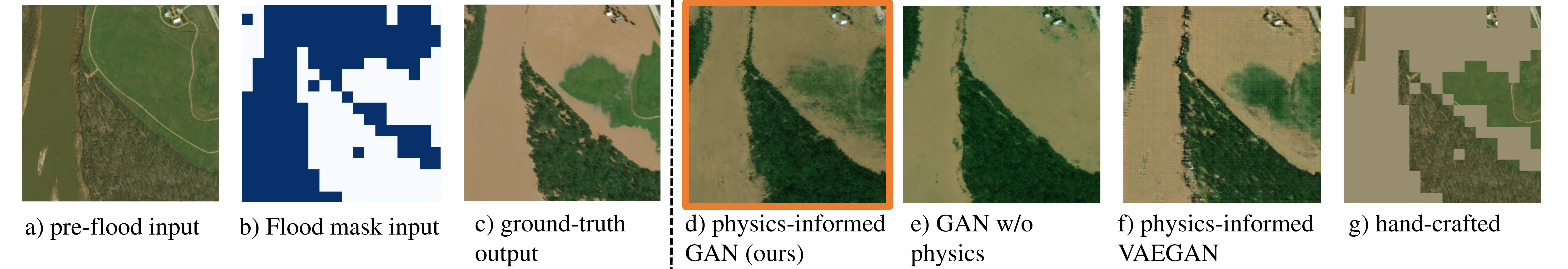}
      \vspace{-.20in}
      %\caption{Before training}
  \end{subfigure}
\caption[Results all]{\textbf{The proposed physics-informed GAN, (d), generates photorealistic and physically-consistent flood imagery from the inputs, (a,b), outperforming all other models, (e,f,g).} The baseline GAN, pix2pixHD~\cite{Wang_2018} (e), in comparison, receives only a pre-flooding image and no physical input. The resulting image, (e), is fully-flooded, rendering the model untrustworthy for emergency response. The VAEGAN, BicycleGAN~\cite{Zhu_2017} (f), creates glitchy imagery (zoom in). A handcrafted baseline model (g), as used in common visualization tools~\cite{ClimateCentral_18, NOAA_2020}, visualizes the correct flood extent, but is pixelated and lacks photorealism.
} 
\label{fig:results_comparison_imagery} 
\vspace{-.1in}
\end{figure}

Recent advances in generative adversarial networks (GANs) generated \textit{photorealistic} imagery of faces~\cite{Isola_2017, Wang_2018}, animals~\cite{Zhu_2017b, brock2018large}, or even  satellite~\cite{requena2019predicting, Fruhstuck_2019}, and street-level flood imagery~\cite{Schmidt_2019}. 
Disaster planners and responders, however, need imagery that is not only photorealistic, but also physically-consistent.
In our implementation, we consider both GANs and variational autoencoders (VAEs), where GANs generate more photorealistic imagery (\cite{Dosovitskiy_2016, Zhu_2017}, \cref{fig:results_comparison_imagery}) and VAEs capture system uncertainties more accurately~\cite{casale2018gaussian, Kingma_2013}. Because our use case requires photorealism to provide intuition, we extend a state-of-the-art, high-resolution GAN, pix2pixHD~\cite{Wang_2018}, to take in physical constraints and produce imagery that is both photorealistic and physically-consistent. We leave ensemble predictions to account for system uncertainties for future work.
% We focus on paired image-to-image translation because for every post-flood image there should exist a pre-flood image. 
 
There are multiple approaches to generating physically-consistent imagery with GANs, where we define \textit{physically-consistent} to assess: \textit{Does the generated imagery depict the same flood extent as the storm surge model?} One approach is conditioning the GAN on the outputs of physics-based models~\cite{Reichstein_2019}; another approach is using a physics-based loss during evaluation~\cite{lesort2019deep}; and yet another is embedding the neural network in a differential equation~\cite{Rackauckas_2020} (e.g., as parameters, dynamics~\cite{Chen_2018}, residual~\cite{Karpatne_2017}, differential operator~\cite{Raissi_2018, Long_2019}, or solution~\cite{Raissi_2019}). Our work focuses on the first two methods, leveraging years of scientific domain knowledge by incorporating a physics-based storm surge model in the image generation and evaluation pipeline.

This work makes three contributions: 1) the first physically-consistent visualization of coastal flood model outputs as high-resolution, photorealistic satellite imagery; 2) a novel metric, the Flood Visualization Plausibility Score (FVPS), to evaluate the photorealism and physical-consistency of generated imagery; and 3) an extensible framework to generate physically-consistent visualizations of climate extremes.

%!TEX root=main.tex
\section{Approach}
The proposed pipeline uses a generative vision model to generate post-flooding images from pre-flooding images and a flood extent map, as shown in~\cref{fig:model_architecture}.

\textbf{Data Overview.} Obtaining post-flood images that display standing water is challenging due to cloud-cover, time of standing flood, satellite revisit rate, and cost of high-resolution imagery. 
This work leverages the xBD dataset~\cite{Gupta_2019}, a collection of pre- and post-disaster images from events like Hurricane Harvey or Florence, from which we obtained ${\sim}3\unit{k}$ pre- and post-flooding image pairs with the following characteristics: ${\sim}.5\unit{m/px}$, RGB, $1024{\times1024}\unit{px/img}$ (post-processing details in~\cref{sec:appendix_dataset}). 
%USE BELOW SENTENCE TO CUT SPACE:
%This work leverages the xBD dataset~\cite{Gupta_2019}, from which we obtained ${\sim}3\unit{k}$, high-resolution (${\sim}.5\unit{m/px}$ and $1024{\times1024}\unit{px/img}$), RGB, pre- and post-flood image pairs from seven flood events (post-processing details in~\cref{sec:appendix_dataset}).
We also downloaded flood hazard maps (at $30\unit{m/px}$), which are outputs of NOAA's widely used storm surge model, SLOSH, that models the dynamics of hurricane winds pushing water on land (\cref{sec:appendix_dataset}). We then aligned the flood hazard map with the flood images and reduced it into a binary flood extent mask (flooded vs. non-flooded).

%(described in more detail in~\cref{sec:appendix_dataset}).
%More precisely, t

%!TEX root=main.tex
%\subsection
\textbf{Model architecture.}
The central model of our pipeline is a generative vision model that learns the physically-conditioned image-to-image transformation from pre-flood image to post-flood image. We leveraged the existing implementation of the GAN pix2pixHD~\cite{Wang_2018} and extended the input dimensions to $1024{\times}1024{\times}4$ to incorporate the flood extent map. Note that the pipeline is modular, such that it can be repurposed for visualizing other climate impacts.

\textbf{The Evaluation Metric Flood Visualization Plausibility Score (FVPS).}
Evaluating imagery generated by a GAN is difficult~\cite{xu2018empirical,Borji_2019}. 
Most evaluation metrics measure photorealism or sample diversity~\cite{Borji_2019}, but not physical consistency~\cite{ravi2019adversarial} (see, e.g., SSIM~\cite{Wang_2004}, MMD~\cite{Bounliphone_2016}, IS~\cite{salimans2016improved}, MS~\cite{Che_2017}
, FID~\cite{heusel2017gans, Zhou_2020}, or LPIPS~\cite{Zhang_2018}). 
%or cycle consistency \cite{Zhu_2017b}
%An extra reason to advocate for physical consistency includes a more accurate approximation of the target distribution.

To evaluate physical consistency, as defined in~\cref{sec:intro}, we propose using the intersection over union (IoU) between water in the generated imagery and water in the flood extent map. This method relies on flood masks, but because there are no publicly available flood segmentation models for RGB imagery, we trained our own model on ${\sim}100$ hand-labeled flooding images (\cref{sec:appendix_dataset}).
This segmentation model produced flood masks of the generated and ground-truth flood image which allowed us to measure the overlap of water in between both.  When the flood masks overlap perfectly, the IoU is 1; when they are completely disjoint, the IoU is 0. 
%To quantify physical consistency, as defined in~\cref{sec:intro}, we had to generate flood masks. As no reliable flood classification model from RBG imagery is publicly available, we trained a segmentation model (based on pix2pix~\cite{Isola_2017} on ${\sim}100$ hand-labelled flooding images) to do pixel-wise flood segmentation from RGB imagery. IoU then measures the overlap of the input flood mask and a generated flood mask of the output imagery (best IoU is $1$, lowest is $0$).

To evaluate photorealism, we  used the state-of-the-art perceptual similarity metric Learned Perceptual Image Patch Similarity (LPIPS)~\cite{Zhang_2018}. LPIPS computes the feature vectors (of an ImageNet-pretrained deep CNN, AlexNet) of the generated and ground-truth tile and returns the mean-squared error between the feature vectors (best LPIPS is $0$, worst is $1$).

Because the joint optimization over two metrics poses a challenging hyperparameter optimization problem, we propose to combine the evaluation of physical consistency (IoU) and photorealism (LPIPS) in a new metric (FVPS), called Flood Visualization Plausibility Score (FVPS). 
The FVPS is the harmonic mean over the submetrics, IoU and $(1-\text{LPIPS})$, that are both $[0,1]$-bounded.
Due to the properties of the harmonic mean, the FVPS is $0$ if any of the submetrics is $0$; the best FVPS is $1$. %TODO: define epsilon
%\blXX{Question (to al, yg, cr): should we use geometric or harmonic mean? The harmonic mean (currently implemented) assumes that the metrics are of the unit and linearly correlated (farrukh), which they are not. The, geometric mean, $\unit{FVPS} = \sqrt{\unit{IoU}^2 \unit{LPIPS}^2}$, would be appropriate for nonlinearly correlated metrics (farrukh).}
\begin{equation}
    \text{FVPS} = \frac{2}{\frac{1}{\text{IoU} + \epsilon} + \frac{1}{1-\text{LPIPS} + \epsilon}}.\label{eq:fvps}
    \vspace{-.1in}
\end{equation} %$GM(\frac{IoU}{L}$ %F.C.: As relayed earlier, I think the geometric mean is appropriate as the two metrics have different units separately, though these are normalized. Harmonic means combines rates of a same metric (such as speeds), while geometric means allows for combination of different metrics} \blXX{todo: rerun evaluation w. geometric mean}
%!TEX root=main.tex
\textbf{}
\section{Experimental Results}
In terms of both physical-consistency and photorealism, our physics-informed GAN outperforms an unconditioned GAN that does not use physics, as well as a handcrafted baseline model (\cref{fig:results_comparison_imagery}).

\textbf{A GAN without physics information generates photorealistic but non physically-consistent imagery.} The inaccurately modeled flood extent in~\cref{fig:results_comparison_imagery}e illustrates the physical-inconsistency and a low IoU of $0.226$ in~\cref{tab:results_comparison_table} over the test set further confirms it (see~\cref{sec:appendix_dataset} for test set details). Despite the photorealism ($\unit{LPIPS}=0.293$), the physical-inconsistency renders the model non-trustworthy for critical decision making, as confirmed by the low FVPS of $0.275$. The model is the default pix2pixHD~\cite{Wang_2018}, which only uses the pre-flood image and no flood mask as input.

\textbf{A handcrafted baseline model generates physically-consistent but not photorealistic imagery.} Similar to common flood visualization tools~\cite{ClimateCentral_18}, the handcrafted model overlays the flood mask input as a hand-picked flood brown (\#998d6f) onto the pre-flood image, as shown in~\cref{fig:results_comparison_imagery}g. Because typical storm surge models output flood masks at low resolution ($30m/px$~\cite{NoaaSlosh_20}), the handcrafted baseline generates pixelated, non-photorealistic imagery. 
Combining the high IoU of $0.361$ and the poor LPIPS of $0.415$, yields a low FVPS score of $0.359$, highlighting the difference to the physics-informed GAN in a single metric.%\blumargin{*}{TODO: re-phrase with brandon bc unclear.}

\begin{figure}[htbp!]%t]%H]
 \centering
 \begin{subfigure}{.8\columnwidth}
  \vspace{-.2in}
  \centering
      \includegraphics [trim=0 .3 0 0, clip, width=1., clip, width=1.\textwidth, angle = 0]{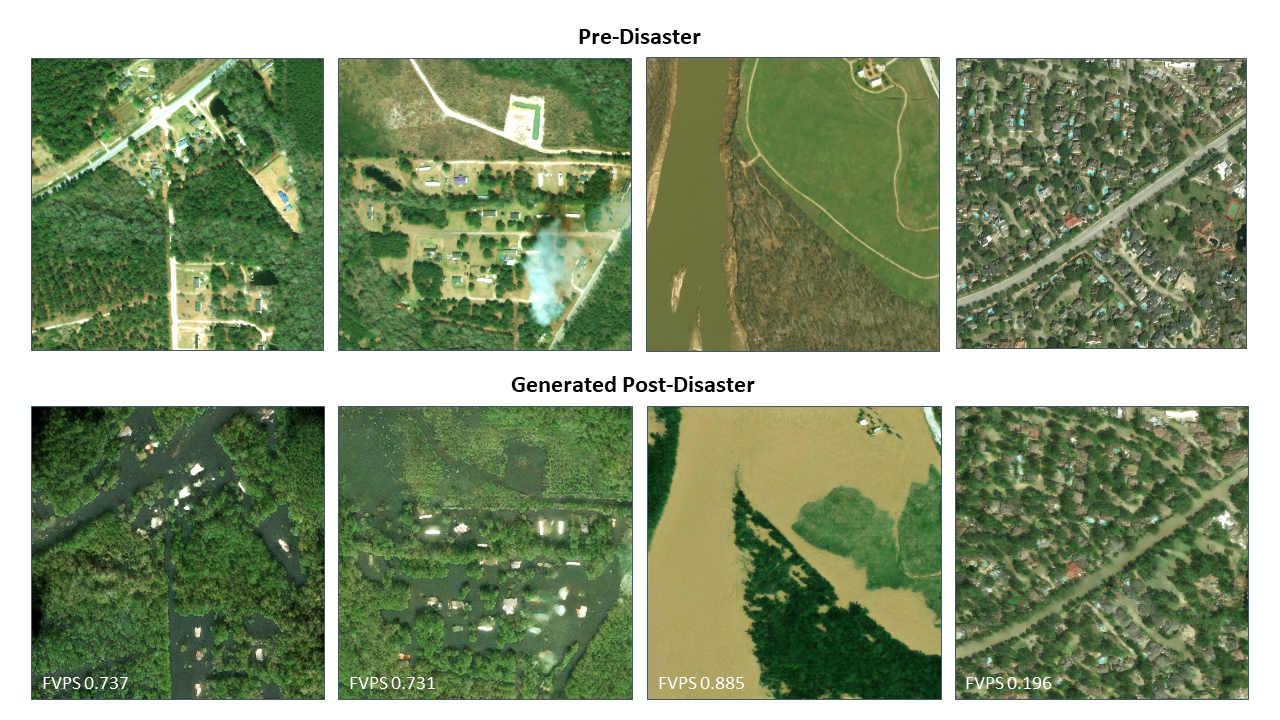}
      \vspace{-.2in}
      %\caption{Before training}
  \end{subfigure}
\caption[teaserfigure]{Our model generates physically-consistent satellite imagery of future coastal flood events to aid in disaster response and climate resilience planning.} \label{fig:results_overview} 
\end{figure}

\textbf{The proposed physics-informed GAN generates physically-consistent and photorealistic imagery.} 
To create the physics-informed GAN, we trained pix2pixHD~\cite{Wang_2018} from scratch on our dataset (${\sim}7\unit{hrs}$ on $8{\times}\unit{V100}$ Google Cloud GPUs). This model successfully learned how to convert a pre-flood image and a flood mask into a photorealistic post-flood image, as shown in~\cref{fig:results_overview}.
% To appendix:
%One training run took ${\sim}7\unit{hrs}$ on $8{\times}\unit{V100}$ Google Cloud GPUs. We also extended the pyTorch implementation~\cite{Wang_2018} to a $4-$ channel input. 
The model outperformed all other models in IoU ($0.553$), LPIPS ($0.263$), and FVPS ($0.532$) (\cref{tab:results_comparison_table}). The learned image transformation %``paints-in``
``in-paints`` the flood mask in the correct flood colors and displays an average flood \textit{height} that does not cover structures (e.g., buildings, trees), as shown in $64$ randomly sampled test images in~\cref{fig:grid_of_gen_ims}. While our model also outperforms the VAEGAN (BicyleGAN), the latter has the potential to create ensemble forecasts over the unmodeled flood impacts, such as the probability of destroyed buildings.%\blumargin{*}{@ndr: is ``destroyed building``, ok?; brandon mentioned that he didnt understand ``probability of buildings showing destroyed``. }

\begin{table*}[tp]
\centering
\vspace{0.1in}
\begin{tabular}{||c||c|c||c|c||c|c||}  
 \hline
  & \thead{LPIPS \\ high res.} & \thead{LPIPS \\ low res.} & \thead{IoU \\ high res.} & \thead{LPIPS \\ low res.}  & \thead{FVPS \\ high res.}  & \thead{FVPS \\ low res.} \\ [0.5ex]
 \hline\hline 
 \textbf{GAN w/ phys. (ours)} & \textbf{0.265} & \textbf{0.283} & \textbf{0.502} & \textbf{0.365} & \textbf{0.533} & \textbf{0.408} \\ \hline
 GAN w/o phys. & 0.293 & \textbf{0.293} & 0.226 & 0.226 & 0.275 & 0.275 \\[0.5ex] \hline
 VAEGAN w/ phys. & 0.449 & - & 0.468 & - & 0.437 & - \\ \hline
 Handcrafted baseline & 0.399 & 0.415 & 0.470  & \textbf{0.361} & 0.411 & 0.359 \\[0.5ex] \hline\hline
 
\end{tabular}
\caption{\textbf{In terms of photorealism (LPIPS) and physical consistency (IoU), our physics-informed GAN outperforms three benchmarks}: the baseline GAN without physics; a physics-informed VAEGAN; and a handcrafted baseline. The proposed Flood Visualization Plausibility Score (FVPS) trades-off IoU and LPIPS as a harmonic mean and highlights the performance differences between the GAN with and without physics on low-resolution flood mask inputs.}
\label{tab:results_comparison_table}
\vspace{-.1in}\end{table*}

%!TEX root=main.tex

\section{Discussion and Future Work}
Although our pipeline outperformed all baselines in the generation of  physically-consistent and photorealistic imagery of coastal floods, there are areas for improvement in future works. For example, our dataset only contained $3k$ samples and is biased towards vegetation-filled satellite imagery; this data limitation likely contributes to our model rendering human-built structures, such as streets and out-of-distribution skyscrapers in~\cref{fig:grid_of_gen_ims} top-left, as smeared. 
%In addition, preliminary results suggest our model does not generalize well to non-Maxar data so
In addition, the dataset was generated by Maxar imagery and preliminary results suggest that our model does not generalize well to other data sources such as NAIP imagery~\cite{NAIP_2019}. 
Although we attempted to overcome our data limitations using several state-of-the-art augmentation techniques, this work would benefit from more public sources of high-resolution satellite imagery (experiment details in \cref{sec:appendix_experiments}).
%As data augmentation showed to be insufficient (\cref{sec:appendix_experiments}) to overcome the data limitations, more sources for public high-resolution satellite imagery and research on transfer learning is needed. 
Finally, the computational intensity of training GANs made it difficult to fine-tune models on new data; improved transfer learning techniques could address this challenge. 
%Despite significant data requirements of GANs, the training also requires significant computational resources which prevented us from improving on NVIDIA's pix2pixHD implementation, as shown in~\cref{sec:appendix_experiments}. 
Lastly, satellite imagery is an internationally trusted source for analyses in deforestation, development, or military~\cite{Hansen_2013, Anderson_2017}, and with the rise of ``deep-fake`` models, more work is needed in the identification of and education around misinformation and ethical AI~\cite{Barredo20}. 
Given our pipeline's results, however, we hope to deploy with NOAA by integrating flood forecasts with aerial imagery along the entire U.S. East Coast.

\textbf{Vision for the future.} 
We envision a global visualization tool for climate impacts.
Our proposed pipeline can generalize in time, space, and type of event. 
By changing the input data, future work can visualize impacts of other well-modeled, climate-attributed events, including arctic sea ice melt, wildfires, or droughts. Non-binary climate impacts, such as inundation height, or drought strength can be generated by replacing the binary flood mask with continuous model predictions. Opportunities are abundant for further work in visualizing our changing Earth, and given its potential impact for both climate mitigation and adaptation, we encourage the ML community to take up this challenge.
%
%to mention in future work journal version:
% Why don you use elevation maps for the water instead of segmentation masks? so this should be argued (relates to acquiring data with SAR right?) clearly even if in appendix if it wasnt already. Alexander Lavin  5:19 PM Yes a good point… when we discussed this we resolved that the information from elevation maps is baked into the segmentation masks. I was suggesting we would want to include the topography values along with the segmentations as multiple “maps” (i.e., multiple constraint matrices), but I believe we didn’t get to those experiments

%%ToDo for us to test: searching for 'physical-consist' I found this work where they use unpaired images with physics constraint loss for different acquisition  settings, may be useful for us to better generalize in future work/geo-areas/disasters
%Adversarial training with cycle consistency for unsupervised super-resolution in endomicroscopy.pdf

\clearpage
% TODO: include in final version, but not in anonymous submission
\subsubsection*{Acknowledgments}
This research was conducted at the Frontier Development Lab (FDL), US. 
The authors gratefully acknowledge support from the MIT Portugal Program, National Aeronautics and Space Administration (NASA), and Google Cloud. 
Further, we greatly appreciate the time, feedback, direction, and help from Prof. Bistra Dilkina, Ritwik Gupta, 
Mark Veillette, Capt. John Radovan, Peter Morales, Esther Wolff, Leah Lovgren, 
Guy Schumann, Freddie Kalaitzis, Richard Strange, James Parr, Sara Jennings, Jodie Hughes, 
Graham Mackintosh, Michael van Pohle, Gail M. Skofronick-Jackson, 
Tsengdar Lee, Madhulika Guhathakurta, Julien  Cornebise, Maria Molina, 
Massy Mascaro, 
Scott Penberthy, Derek Loftis, Sagy Cohen, 
John Karcz, Jack Kaye, 
Janot Mendler de Suarez, Campbell Watson, and all other FDL researchers. 

Björn Lütjens and Brandon Leshchinskiy's research has been partially sponsored by the United States Air Force Research Laboratory and the United States Air Force Artificial Intelligence Accelerator and was accomplished under Cooperative Agreement Number FA8750-19-2-1000. The views and conclusions contained in this document are those of the authors and should not be interpreted as representing the official policies, either expressed or implied, of the United States Air Force or the U.S. Government. The U.S. Government is authorized to reproduce and distribute reprints for Government purposes notwithstanding any copyright notation herein.

\iffalse
\subsubsection*{Author Contributions}
\fi

\bibliographystyle{IEEEtran}
\bibliography{references}

\newpage
%!TEX root=main.tex
\section{Appendix}
\subsection{Additional Results}
\begin{figure}[h]%tbp!]%t]
 \centering
 \begin{subfigure}{1.\columnwidth}
  \centering
      \includegraphics [trim=0 0 0 0, clip, width=1.\textwidth, angle = 0]{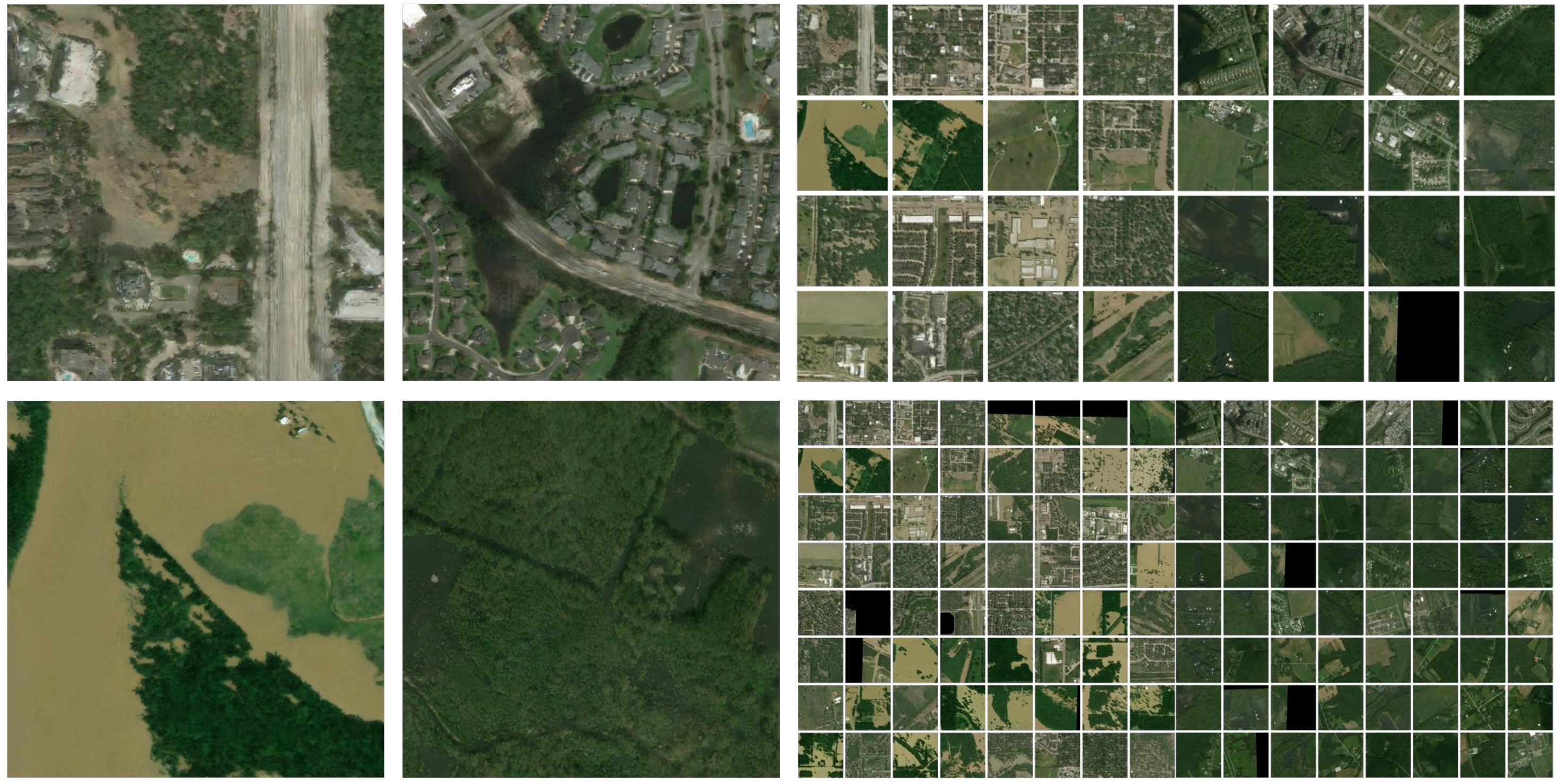}
      \vspace{.02in}
      %\caption{Before training}
  \end{subfigure}
\caption[Unconditioned model]{\textbf{Generated post-flooding imagery of $64$ randomly chosen tiles} of hurricanes Harvey and Florence test set.} \label{fig:grid_of_gen_ims} 
\end{figure}
\cref{fig:grid_of_gen_ims} shows an additional set of randomly sampled generated imagery from the test set.~\Cref{fig:grid_of_gen_ims} shows $4$ additional high-resolution test pictures with the corresponding pre-flood image.

\subsection{Dataset}\label{sec:appendix_dataset}
\subsubsection{Pre- and post-flood imagery}
Post-flood images that display standing water are challenging to acquire due to cloud-cover, time of standing flood, satellite revisit rate, and cost of high-resolution imagery. To the extent of the authors' knowledge, xBD~\cite{Gupta_2019} is the best publicly available data-source for preprocessed high-resolution imagery of pre- and post-flood images. More open-source, high-resolution, pre- and post-disaster images can be found in unprocessed format on DigitalGlobe's Open Data repository~\cite{Digitalglobe_2020}.
\begin{itemize}
    \item Data Overview: $3284$ flood-related RGB image pairs from seven flood events at $1024{\times}1024\unit{px}$ of ${\sim}0.5\unit{m/px}$ resolution of which 30\% display a standing flood (${\sim}1370$).
    \item Flood-related events: hurricanes (Harvey, Florence, Michael, Matthew in the U.S. East or Golf Coast), spring floods (Midwest U.S., ‘19), tsunami (Indonesia), monsoon (Nepal).
    \item Our evaluation test set is composed by 108 images of each hurricane Harvey and  Florence. The test set excludes imagery from hurricane Michael or Matthew, because the majority of tiles does not display standing flood. % 98 of Hurricane Michael, and 73 of hurricane Matthew. TODO: add nepal and midwest to test set!!!
    \item We did not used digital elevation maps (DEMs), because the information of low-resolution DEMs is contained in the storm surge model and high-resolution DEMs for the full U.S. East Coast are not publicly available.
\end{itemize}
An important part of pre-processing the xBD data was to correct the geospatial references per image. Correcting the geolocation is necessary to extrapolate our model to visualize future floods across the full U.S. East Coast, based on storm surge model outputs~\cite{NOAA_2020} and high-resolution imagery~\cite{NAIP_2019}. To align the imagery, we (1) extracted tiles from NAIP that approximately match xBD tiles via google earth engine, (2) detected keypoints in both tiles via AKAZE, (3) identified matching keypoints via l2-norm in image coordinates, (4) approximated the homography matrix between two feature matrices via RANSAC, and (5) applied the homography matrix to transform the xBD tile.

\subsubsection{Flood segmentations}
For $111$ post-flood images, segmentation masks of flooded/non-flooded pixels were manually annotated to train a pix2pix segmentation model~\cite{Isola_2017} from scratch. The model consisted of a vanilla U-Net for the generator that was trained with L1-loss, IoU, and adversarial loss; its last layers were finetuned solely on L1-loss. A four-fold cross validation was performed leaving $23$ images for testing. The segmentation model selected to %become part of 
be used by the FVPS has a mean IoU performance of $0.343$.
Labelled imagery will be made available at the project GitLab. 
%We extended the PyTorch implementation of Pix2pixHD ~\cite{Wang_2018} to a $4-$ channel input, and trained from scratch on our dataset. One training run of Pix2pixHD~\cite{Wang_2018} took ${\sim}7\unit{hrs}$ on $8{\times}\unit{V100}$ Google Cloud GPUs. 

\subsubsection{Storm Surge predictions}
Developed by the National Weather Service (NWS), the Sea, Lake and Overland Surges from Hurricanes (SLOSH) model~\cite{Jelesnianski_92} estimates storm surge heights from atmospheric pressure, size, forward speed and track data, which are used as a wind model driving the storm surge. The SLOSH model consists of shallow water equations, which consider unique geographic locations, features and geometries. The model is run in deterministic, probabilistic and composite modes by various agencies for different purposes, including NOAA, National Hurricane Center (NHC) and NWS. We use outputs from the composite approach -- that is, running the model several thousand times with hypothetical hurricanes under different storm conditions. As a result, we obtain a flood hazard map as displayed in~\cref{fig:related_works}a which are storm-surge, height-differentiated, flood extents. Future works will use the state-of-the-art ADvanced CIRCulation model (ADCIRC)~\cite{Luettich_1992} model, which has a stronger physical foundation, better accuracy, and higher resolution than SLOSH. ADCIRC storm surge model output data is available for the USA from the Flood Factor online tool developed by First Street Foundation.

\subsection{Experiments}\label{sec:appendix_experiments}

Standard data augmentation, here rotation, random cropping, hue, and contrast variation, and state-of-the art augmentation - here elastic transformations~\cite{simard2003best} - were applied. Further, spectral normalization~\cite{miyato2018spectral} was used to stabilize the training of the discriminator. And a relativistic loss function has been implemented to stabilize adversarial training. We also experimented with training pix2pixHD on LPIPS loss. Quantitative evaluation of these experiments, however, showed that they did not have significant impact on the performance and, ultimately, the results in the paper have been generated by the pytorch implementation~\cite{Wang_2018} extended to $4$-channel inputs. %Todo ablation study to be shown in journal?

\textbf{Pre-training LPIPS on satellite imagery.} The standard LPIPS did not clearly distinguish in between the handcrafted baseline and the phyiscs-informed GAN, contrasting the opinion of a human evaluator. This is most likely because LPIPS currently leverages a neural network that was trained on object classification from ImageNet. The neural network might not be capable to extract meaningful high-level features to compare the similarity of satellite images. In preliminary tests the ImageNet-network would classify all satellite imagery as background image, indicating that the network did not learn features to distinguish satellite imagery. Future work, will use LPIPS with a network trained to have satellite imagery specific features, e.g., Tile2Vec or a land-use segmentation~\cite{Robinson_2019} model. %Advanced: Modified FID with a feature space that is pretrained on a satellite imagery segmentation task,extending on~\cite{Zhou_2020}}
% for journal: add thoughts on why high-res is necessary; add thought on climate change isn't v. visible}

\subsection{Further discussion: Ethical Implications.} 

Satellite imagery is an internationally trusted data source to conduct analyses in deforestation, development, logistics, or military~\cite{Hansen_2013, Anderson_2017}. Generating artificial satellite imagery can enable various stakeholders with malicious intent to, e.g., depict fake military operations, and result in a loss of trust in satellite imagery. Hence, we have put a strong focus onto generating physically-consistent imagery and clearly label our imagery as artificial, following the guidelines for responsible AI \cite{Barredo20}. We further encourage analyses to source data from trusted sources (e.g., NASA, ESA, or PT Space) and public education around misinformation and ethical AI.

% \section*{References}

% \medskip

% \small

\end{document}